\documentclass[11pt]{article}
\usepackage[preprint]{acl}
\usepackage{latexsym}
\usepackage{graphicx}
\usepackage{booktabs}
\usepackage{amssymb}
\usepackage{amsmath}
\usepackage{enumitem}
\usepackage{placeins}
\usepackage{microtype}
\usepackage[english,bidi=default]{babel}
\babelfont{rm}[Extension=.otf, UprightFont=*-Regular, BoldFont=*-Bold, ItalicFont=*-Italic,
  BoldItalicFont=*-BoldItalic]{TeXGyreTermesX}          
\babelprovide[import]{arabic}
\babelfont[*arabic]{rm}[Extension=.ttf, UprightFont=*-Regular,
  BoldFont=*-Bold]{Amiri}          
\newcommand{\textarabic}[1]{\foreignlanguage{arabic}{#1}}

\title{What the Query Remembers: Query Abstraction Governs\\Lexical and Dense Retrieval of Classical Arabic Poetry}

\author{
  Omar Farouk Zouak\thanks{\ Primary contributor to this work.} \\
  National School of Artificial Intelligence \\
  \texttt{omar.zouak@ensia.edu.dz}
  \And
  Selma Mani \\
  National School of Artificial Intelligence \\
  \texttt{selma.mani@ensia.edu.dz}
  \AND
  Larbi Said Chikh \\
  National School of Artificial Intelligence \\
  \texttt{larbi.saidchikh@ensia.edu.dz}
  \And
  Raid Abderrezak Ouahioune \\
  National School of Artificial Intelligence \\
  \texttt{raid.ouahioune@ensia.edu.dz}
}

\begin{document}
\maketitle

\begin{abstract}
How well a poetry archive can be searched depends on what the user still remembers, not only on what the model understands. We show this for Classical Arabic poetry, benchmarking nine dense encoders and BM25 over two corpora, three indexing modes and two evaluation protocols. Which retriever wins is governed by how much of a poem's wording the query still carries. On tip-of-the-tongue queries BM25 leads every dense encoder, including a large multilingual embedding model (Qwen3-Embedding-8B, 0.746 vs.\ 0.717 Recall@5); on abstract queries it falls to eighth out of ten (0.392 vs.\ 0.748). Editing the wording of the same queries shows that this ranking largely reflects query composition: adding one remembered half-verse raises BM25 by 38 points, rewriting the quoted words by hand leaves BM25 level with the 8B encoder, and BM25's sensitivity to remembered wording is significantly steeper than that of the strongest dense encoder. The same axis controls explanation augmentation, whose effect swings from $+37$ to $-39$ points. The mechanism splits by system type: lexical containment best predicts how BM25 responds to explanations, embedding proximity how dense encoders do. Finally, the field's two standard protocols, known-item retrieval and thematic relevance, dissociate on the same systems, a result reproduced by two blind human annotators and by two open-weights judges, one of them under a rubric with no literal-match criterion. A benchmark that reports one protocol, or one query register, can give a materially wrong picture. Code and data are available at \url{https://github.com/Ellzo/what-the-query-remembers}.
\end{abstract}

\section{Introduction}
\label{sec:intro}

Classical Arabic poetry, from the pre-Islamic odes to the Abbasid and Andalusian eras, is a central pillar of Arab cultural heritage and a uniquely rich linguistic archive. It also poses distinctive problems for retrieval. Root-and-pattern morphology yields dense variation; diacritics are usually omitted; and rhetorical devices such as \textit{tawriya} (double meaning) and \textit{jin\=as} (paronomasia) fuse literal and symbolic meaning inside the rigid \textit{bayt} (verse) structure. The surface form of a verse is a poor guide to its meaning, which makes poetry an attractive stress test for embedding models that claim to retrieve by meaning rather than wording.

Two gaps motivate this work. The first is the domain: work on Arabic poetry centres on \emph{classification and generation} (meter, authorship, composition), while \emph{retrieval} is largely unstudied, despite underpinning digital archives, education and literary scholarship. The second is more general. Lexical and dense retrievers are usually compared on a fixed query set, so how the queries themselves shape the comparison is seldom measured directly. We asked how well modern embedding models search Classical Arabic poetry by meaning, and found that the answer depends on the queries as much as on the models.

\paragraph{The finding.} Users arrive with different amounts of remembered text: some half-remember a hemistich, others recall only a theme. We place queries on a \emph{query-abstraction spectrum}, where abstraction means how little of the target poem's own wording a query carries, and we measure it by surface overlap with the poem. We call a query \emph{quoting} when it shares a verbatim span of three or more words with its poem. A query's position along this spectrum strongly influences the relative effectiveness of lexical and dense retrieval, and whether a standard improvement technique helps or hurts.

On tip-of-the-tongue queries, a plain BM25 index beats all nine dense encoders, including Qwen3-Embedding-8B ($0.746$ vs.\ $0.717$ Recall@5); on abstract queries it falls to eighth of ten ($0.392$ vs.\ $0.748$). Editing the same queries shows that this reversal largely reflects the wording they carry. Appending one verbatim half-verse of the target poem raises BM25 by $37.8$ points, against $12.8$ to $23.2$ for the 8B encoder, and rewriting the quoted words by hand brings BM25 down to the 8B encoder's level.

The same axis governs \emph{explanation augmentation}, the natural remedy for figurative text: indexing a short interpretive gloss of each verse instead of the verse. Compared with indexing the verses, it changes Recall@5 by up to $+37$ points where queries are abstract and as much as $-39$ where they carry remembered wording, and a purely lexical retriever shows the largest swing in both directions.

The queries also decide how a system fares under the field's two standard evaluation protocols. Known-item retrieval and thematic relevance disagree on the same systems: on the tip-of-the-tongue corpus, BM25 is first of ten under one and last under the other, and blind human judgments reproduce the disagreement rather than resolving it. The effect sizes are specific to our corpora, but measuring overlap and editing queries apply to any known-item benchmark.

\paragraph{Contributions.}
\begin{enumerate}[noitemsep,topsep=2pt,leftmargin=1.4em]
\item To our knowledge, the first systematic study of retrieval for Classical Arabic poetry: nine dense encoders and a lexical baseline, two corpora, three indexing modes, two evaluation protocols, all from one reproducible pipeline, with a released query benchmark of $502$ known-item and $200$ thematic queries.
\item The query-abstraction spectrum as an explanatory variable, established by editing the wording of the same queries and by a within-corpus stratification rather than by a cross-corpus correlation, together with a graded account of the mechanism: surface containment best predicts the lexical retriever's response to explanations, embedding proximity that of dense encoders.
\item Evidence that the two standard evaluation protocols dissociate, validated by two blind human annotators and three LLM judges from two model families, and a deployment recipe whose correct form depends on the query register a system will serve.
\end{enumerate}

\section{Related Work}
\label{sec:related}

Most Arabic encoders \citep{antoun-etal-2020-arabert,abdul-mageed-etal-2021-arbert,inoue-etal-2021-interplay} are trained on prose, news or dialect. Poetry-specific work concentrates on classification and generation: authorship attribution \citep{alqurashi-etal-2025-bert}, meter classification \citep{mutawa2025meter}, rhythm-aware composition \citep{elzohbi2025tahdib}, and resources such as the \textit{Diwan} corpus \citep{alonazi2025diwan}. AraPoemBERT \citep{qarah2024arapoembert} targets poetic understanding but not retrieval, and recent taxonomies of poetry NLP list retrieval as a distinct, unaddressed capability \citep{chalamalasetti2025metricalargs}.

On the dense retrieval side, general-purpose multilingual encoders, including gte \citep{li2023gte,zhang2024mgte}, jina-v3 \citep{sturua2024jina}, LaBSE \citep{feng-etal-2022-labse}, BGE-M3 \citep{chen2024bgem3}, multilingual-E5 \citep{wang2024multilinguale5} and Qwen3-Embedding \citep{zhang2025qwen3embedding}, alongside Arabic-oriented models \citep{silma2024embedding,muffakir2025embedding}, are tuned on similarity data dominated by prose. Arabic-centric benchmarks such as ArabicMTEB \citep{bhatia2024swan} broaden dialect and task coverage, but figurative classical verse is not their focus. \textit{Fann or Flop} \citep{alghallabi2025fannorflop} supplies reference explanations of poems but frames the task as comprehension, not retrieval.

That lexical and dense retrievers have complementary failure modes is established in mainstream IR. BEIR \citep{thakur2021beir} found BM25 a strong out-of-domain baseline that many dense models fail to beat, and \citet{sciavolino-etal-2021-simple} showed dense retrievers underperforming BM25 precisely on entity-centric questions, where success requires matching a specific surface form. Analyses of the representational limits of fixed-width dense encoders \citep{luan-etal-2021-sparse} and attempts to give dense retrievers lexical behaviour \citep{chen2021spar} run along the same axis and motivate the sparse-dense hybrids we evaluate in \S\ref{sec:recipe}. We do not claim to have found the axis. What we add is evidence from editing the queries themselves, within one corpus, rather than from comparing datasets that differ in many ways at once; the finding that the same axis flips the sign of document expansion and makes two standard evaluation protocols dissociate; and a measurement where the gap between query and document wording is widest by construction, in figurative verse rather than factoid text. Because this literature documents the effect on naturally occurring queries, our query sets reproduce a known phenomenon rather than manufacture one.

A complementary line of work addresses the gap between queries and documents through expansion. Indexing generated text alongside or instead of a document is a well-established retrieval device: doc2query predicts queries a document should answer and appends them before indexing \citep{nogueira2019doc2query}, while HyDE expands the opposite side, generating a hypothetical document from the query \citep{gao2022hyde}. Explanation augmentation is a form of document expansion, so the technique itself is not new. Our question is \emph{when and why} it works: whether it helps or hurts, and by how much, depends on the query register. Most expansion work uses web and open-domain collections with keyword-like queries, where this dependence is hard to observe.

Closely related is the problem of known-item and tip-of-the-tongue retrieval, where a user searches for a specific half-remembered item. This is a recognised task with its own evaluation tradition, formalised in the ToT known-item setting \citep{arguello2021tot} and the TREC Tip-of-the-Tongue track \citep{arguello2023trectot}. Our known-item queries are deliberately of this kind, so remembered wording is a property of the task, not a defect of the benchmark, and it is what we manipulate in \S\ref{sec:bridge}.

Finally, our thematic evaluation relies on LLM-based judgment, standard for scoring open-ended output but with reliability caveats that depend on judge capacity and rubric design \citep{zheng2023judging}. We treat the judge as a hypothesis to be tested rather than a ground truth, validating it against blind human judgments and two open-weights judges (\S\ref{sec:trackb}).

\section{Experimental Setup}
\label{sec:data}

\subsection{Task Definition}
\label{sec:task}

Given a natural-language query in Arabic, the task is to retrieve the poem that best matches it. Each poem is segmented into overlapping three-verse chunks. Each chunk is embedded independently, and chunk scores are max-pooled to the poem level.

\subsection{Data}
\label{sec:datasubsec}

We use two corpora. \textit{Fann or Flop} \citep{alghallabi2025fannorflop} provides poems together with reference explanations, described by its authors as human-written; we refer to these as \emph{FoF explanations} and make no independent claim about their authorship. We keep the $4{,}687$ of its $6{,}984$ poems that have at most 16 verses: longer poems were costly to explain and embed, would dominate a max-pooled index, and several had repetitive verses or explanations. \textit{Diwan} \citep{alonazi2025diwan} is a 14M-verse corpus; for tractability we sample 70 poems from each of the 50 most frequent poets ($3{,}500$ poems). For both corpora we generate a short AI explanation of every verse with Gemini 2.0 Flash, using a fixed instruction that asks for a faithful prose paraphrase and forbids adding information beyond the verse. A manual audit of $50$ of them found that $8\%$ misread the verse (Appendix~\ref{app:corpus}).

We compare three indexing modes: verses (raw verse text, the baseline), AI explanations, and FoF explanations (\textit{Fann or Flop} only), where the interpretive text replaces the verse before embedding. All explanations are generated independently of the queries, and no retriever sees a query at indexing time.

The benchmark contains two known-item query sets, written by the authors under deliberately different instructions, and one thematic set (Table~\ref{tab:data}). Together these define the \emph{query-abstraction spectrum}, the paper's independent variable. \textit{Fann or Flop} queries simulate a user who \emph{half-remembers the poem}: writers saw the poem and were free to include remembered wording, as a real tip-of-the-tongue searcher would \citep{arguello2021tot}. \textit{Diwan} queries were deliberately written as \emph{abstract descriptions} of what a poem is about, with little quotation. The thematic queries name a topic only (``pride and valour'') and were written before the corpora were inspected, so they have no target poem at all.

\paragraph{Provenance.} Both known-item sets were written from the poem text alone. Writers saw only the verses: they did not see the FoF reference explanations, and the AI explanations did not exist until after the queries were written. The two instructions differed in register and in nothing else. The target poems do not differ from the rest of their corpus in era, genre or length (all $p>.05$; Appendix~\ref{app:corpus}).

We quantify the spectrum with two statistics per query: the longest verbatim token $n$-gram shared with the target poem, and lexical containment $C(Q,D)=|\mathrm{tok}(Q)\cap\mathrm{tok}(D)|/|\mathrm{tok}(Q)|$ on diacritic-stripped normalised tokens. $32.8\%$ of \textit{Fann or Flop} queries share a verbatim three-gram with their target, against $2.7\%$ for \textit{Diwan}, and mean containment differs correspondingly ($0.368$ vs.\ $0.151$). Representative queries, one per register:

\begin{itemize}\setlength\itemsep{1pt}\small
  \item \textbf{ToT (quoting):} \textarabic{قصيدة هجاء في حبيبة قبيحة مطلعها أكثري أو فأقلي قد مللناك فملي} \\ \textit{(``A satirical poem about an ugly beloved, opening \emph{Say more or say less, we have tired of you, so desist}'')} the opening hemistich is quoted verbatim.
  \item \textbf{Abstract:} \textarabic{قصيدة فلسفية ترفض فكرة تناسخ الأرواح وتنصح بعدم تصديق ما لا يؤيده العقل} \\ \textit{(``A philosophical poem rejecting the transmigration of souls, counselling trust only in what reason supports'')} only two isolated words are shared with the poem.
  \item \textbf{Thematic:} \textarabic{الشيب وأثره على نفسية الإنسان} \\ \textit{(``Greying hair and its effect on the psyche'')} no target poem exists.
\end{itemize}

\begin{table}[t]
\centering\small
\setlength{\tabcolsep}{4pt}
\begin{tabular}{lrrrrrr}
\toprule
Query set & $n$ & len & \%\,quote & $C_{\mathrm{vrs}}$ & $C_{\mathrm{ai}}$ & $C_{\mathrm{ff}}$ \\
\midrule
FannOrFlop & 244 & 10.5 & 32.8 & 0.368 & 0.254 & 0.461 \\
Diwan & 258 & 12.8 & 2.7 & 0.151 & 0.360 & n/a \\
\bottomrule
\end{tabular}
\caption{The query-abstraction spectrum. \%\,quote is the share of queries sharing a verbatim three-token span with the target poem; $C$ is mean query-document containment against verses, AI explanations and FoF explanations. The FoF explanations are almost four times longer than the verses and reuse about half of each poem's vocabulary, which raises $C_{\mathrm{ff}}$. The AI explanations reuse about a fifth, so they lower overlap for queries that quote the poem and raise it for queries written as prose descriptions. Thematic queries ($n{=}200$) have no target poem and are used only in Track~B.}
\label{tab:data}
\end{table}

\subsection{Evaluation}
\label{sec:eval}

We use two evaluation protocols. Track A is known-item retrieval: each query has one gold poem, and we report Recall@1, Recall@5, MRR and nDCG@10 with bootstrap 95\% confidence intervals ($10^4$ resamples), comparing systems and indexing modes with paired exact McNemar tests. It needs no judge. Track B is thematic search on the 200 topical queries, which have no single correct poem. An LLM judge scores each system's results on seven aspects from 1 to 5, and we check it against blind human judgments and two open-weights judges (\S\ref{sec:trackb}). Because we run many tests, we apply Holm correction within each group of related comparisons (Appendix~\ref{app:mult}); three contrasts that are significant only before correction are shown unstarred in Table~\ref{tab:main}.

All systems, corpora, modes and query sets run through a single pipeline that dumps per-query ranked lists, so no table mixes sources. Dense encoders receive diacritic and tatweel stripping only; BM25 additionally receives the orthographic unification standard for Arabic lexical matching. This split is deliberate and measured. Stripping diacritics is essential for all three encoders we tested ($+27$ to $+47$ Recall@5 points); the further unification of \textit{alef}, \textit{ya} and \textit{ta-marbuta} hurts all three but varies fourfold in size, from $1.9$ to $8.9$ points (Appendix~\ref{app:norm}). Each encoder uses the query/passage convention prescribed by its own model card.

\section{Track A: Known-Item Retrieval}
\label{sec:tracka}

In Track A, which system ranks first depends on the query set (Table~\ref{tab:main}). On \textit{Fann or Flop}, BM25 ranks above every dense encoder, including Qwen3-8B ($0.746$ vs.\ $0.717$ Recall@5, a gap that is not itself significant, $p{=}.47$). On \textit{Diwan} the ordering reverses: BM25 falls to eighth of ten and Qwen3-8B leads it by 36 points ($p<10^{-20}$). Corpus and query register change together between these two columns, so this comparison alone cannot say which of them is responsible; \S\ref{sec:bridge} separates them within one corpus. Scaling within one family from Qwen3-0.6B to Qwen3-8B adds $17$ and $22$ points respectively, yet the 8B model still does not surpass BM25 on \textit{Fann or Flop}: capacity alone does not close the gap.

This query-dependence extends to explanation augmentation. On \textit{Diwan}, replacing verses with AI explanations improves every system, by up to $+37$ points (BM25 $0.392\!\rightarrow\!0.760$; all ten gains significant at $p<10^{-4}$). On \textit{Fann or Flop} the same substitution hurts most systems, most severely BM25 ($0.746\!\rightarrow\!0.357$, $-39$ points). The richer FoF explanations behave differently again, helping on \textit{Fann or Flop} where the AI explanations hurt.

\begin{table*}[t]
\centering\small
\begin{tabular}{l ccc c cc}
\toprule
& \multicolumn{3}{c}{\textit{FannOrFlop} (tip-of-the-tongue)} & & \multicolumn{2}{c}{\textit{Diwan} (abstract)} \\
\cmidrule(lr){2-4}\cmidrule(lr){6-7}
System & verses & AI expl. & FoF expl. & & verses & AI expl. \\
\midrule
Qwen3-8B & 0.717 & 0.656 & 0.816$^{*}$ & & 0.748 & 0.930$^{***}$ \\
BGE-M3 & 0.693 & 0.598 & 0.725 & & 0.574 & 0.845$^{***}$ \\
mE5-large & 0.578 & 0.492 & 0.717$^{**}$ & & 0.585 & 0.802$^{***}$ \\
jina-v3 & 0.647 & 0.504$^{**}$ & 0.676 & & 0.523 & 0.756$^{***}$ \\
silma-sts & 0.566 & 0.508 & 0.717$^{***}$ & & 0.558 & 0.810$^{***}$ \\
gte & 0.561 & 0.389$^{***}$ & 0.635 & & 0.488 & 0.806$^{***}$ \\
Qwen3-0.6B & 0.549 & 0.430$^{*}$ & 0.660$^{*}$ & & 0.527 & 0.795$^{***}$ \\
LaBSE & 0.311 & 0.283 & 0.336 & & 0.368 & 0.670$^{***}$ \\
Muffakir-v2 & 0.209 & 0.287 & 0.467$^{***}$ & & 0.233 & 0.585$^{***}$ \\
\midrule
\textbf{BM25} & 0.746 & 0.357$^{***}$ & 0.742 & & 0.392 & 0.760$^{***}$ \\
\bottomrule
\end{tabular}
\caption{Track A, Recall@5, one pipeline throughout. Stars mark a significant change from that system's verse baseline (paired exact McNemar, Holm-corrected across the ten systems within each corpus and contrast): $^{*}p{<}.05$, $^{**}p{<}.01$, $^{***}p{<}.001$. Confidence intervals in Appendix~\ref{app:full}. BM25 ranks first on \textit{Fann or Flop} and eighth on \textit{Diwan}; augmentation reverses sign between the two.}
\label{tab:main}
\end{table*}

\section{The Query-Abstraction Effect}
\label{sec:bridge}

The pattern in \S\ref{sec:tracka} suggests a hypothesis. A query succeeds against a figurative verse to the extent that some bridge exists between them: remembered wording supplies one directly, an interpretive gloss indirectly, by restating the verse in the prose register queries use. If so, retrieval should track the amount of bridge available, the sign of augmentation should follow whether a gloss adds or removes it, and a retriever with no notion of meaning should show the effect most strongly. We test this in two steps within \textit{Fann or Flop}: first by stratifying its existing queries, which shows an association, and then by editing the wording of the same queries, which tests the effect directly.

\subsection{Stratifying the existing queries}
\label{sec:strat}

We split the 244 \textit{Fann or Flop} queries by whether they share a verbatim three-token span with the target poem (80 do, 164 do not) and re-evaluate every system on each subset (Table~\ref{tab:controlled}). BM25 retrieves the gold poem within the top five for all $80$ quoting queries and drops to $0.622$ without quotation, a swing of $37.8$ points. For BM25 this is almost true by construction, since the split is defined by shared wording. What we compare across systems is the drop from the quoting to the non-quoting subset, which we call the \emph{slope}. Dense encoders do not reach the ceiling on quoting queries (the best, BGE-M3, reaches $0.875$).

Every system loses accuracy when quotation is removed ($13.5$ to $37.8$ points), so the axis is real for dense encoders too; but BM25's loss is the largest of the ten, and significantly larger than the strongest dense encoder's (difference-in-differences $+16.2$ points, paired bootstrap over $10^4$ resamples, uncorrected $p{=}.005$, Holm-corrected across the nine comparisons $p{=}.045$). No other system differs significantly from Qwen3-8B on this measure. Our claim concerns this interaction rather than either main effect. BM25's slope is also the steepest of the ten under Recall@1, MRR and nDCG@10, with differences of $+0.18$, $+0.16$ and $+0.14$ against Qwen3-8B (uncorrected $p{=}.009$, $.006$, $.009$; Holm-corrected $.085$, $.050$, $.079$; Appendix~\ref{app:robust}).

On this split, \textit{Fann or Flop}'s aggregate ranking reflects its query mix. BM25's overall lead comes entirely from the quoting subset: it beats Qwen3-8B there ($1.000$ vs.\ $0.863$, exact McNemar $p{=}.001$), while on the $164$ non-quoting queries the two are statistically indistinguishable ($0.622$ vs.\ $0.646$, $p{=}.69$), as are their aggregate scores ($0.746$ vs.\ $0.717$, $p{=}.47$). Repeating the split at spans of two to five tokens leaves BM25's slope significantly steeper than Qwen3-8B's throughout ($p$ between $.0001$ and $.049$; Appendix~\ref{app:threshold}).

\subsection{Editing the same queries}
\label{sec:edits}

A stratification compares different queries, which may also differ in specificity or in the difficulty of their target poems. We therefore change the wording of the same queries while keeping each information need and target poem fixed (Table~\ref{tab:edits}; construction details in Appendix~\ref{app:edits}).

\paragraph{Adding wording.} To each of the 164 non-quoting queries we append one verbatim half-verse of its target poem, taken from the opening, a middle or the final verse. BM25 gains $37.8$ points at every position and reaches $1.000$, so its gain is capped. Qwen3-8B gains $23.2$, $20.7$ and $12.8$ points, and the difference is significant at every position ($p\leq.004$). Dense encoders gain most from opening half-verses and least from final ones; BM25 gains the same wherever the half-verse comes from. Only gte, at the opening position, gains slightly more than BM25.

\paragraph{Removing wording.} A co-author rewrote by hand the quoted words of 79 quoting queries, keeping the information need and any proper names; mean query length is preserved ($10.9$ to $11.3$ tokens). BM25 falls from $1.000$ to $0.646$ and Qwen3-8B from $0.861$ to $0.671$, a significant difference ($-0.165$, $p{=}.007$) that leaves the two level, as on the non-quoting subset. Rewrites produced by an LLM and checked by a second co-author agree ($55$ valid rewrites, $p{=}.006$). One system falls further than BM25: gte drops from $0.785$ to $0.405$, so BM25 is the most wording-dependent system on average but not uniquely so.

\begin{table}[t]
\centering\footnotesize
\setlength{\tabcolsep}{2.5pt}
\begin{tabular}{l ccc c}
\toprule
& \multicolumn{3}{c}{half-verse added, $n{=}164$} & rewritten \\
\cmidrule(lr){2-4}
System & opening & middle & final & $n{=}79$ \\
\midrule
Qwen3-8B & ${+}.232$ & ${+}.207$ & ${+}.128$ & ${-}.190$ \\
BGE-M3 & ${+}.287$ & ${+}.207$ & ${+}.055$ & ${-}.266$ \\
mE5-large & ${+}.348$ & ${+}.183$ & ${+}.079$ & ${-}.203$ \\
jina-v3 & ${+}.262$ & ${+}.140$ & ${-}.018$ & ${-}.253$ \\
silma-sts & ${+}.274$ & ${+}.256$ & ${+}.146$ & ${-}.139$ \\
gte & ${+}.384$ & ${+}.238$ & ${+}.104$ & ${-}.380$ \\
Qwen3-0.6B & ${+}.329$ & ${+}.201$ & ${+}.079$ & ${-}.228$ \\
LaBSE & ${+}.152$ & ${+}.116$ & ${+}.030$ & ${-}.139$ \\
Muffakir-v2 & ${+}.128$ & ${+}.085$ & ${+}.024$ & ${-}.038$ \\
\midrule
\textbf{BM25} & ${+}.378$ & ${+}.378$ & ${+}.378$ & ${-}.354$ \\
\midrule
difference & ${+}.146^{**}$ & ${+}.171^{***}$ & ${+}.250^{***}$ & ${-}.165^{**}$ \\
\bottomrule
\end{tabular}
\caption{Change in Recall@5 on \textit{Fann or Flop} when the same queries are edited. Left: one verbatim half-verse of the target poem appended to each non-quoting query, taken from the opening, a middle or the final verse. Right: the quoted words of the quoting queries rewritten by hand at equal length. Last row: BM25's change minus Qwen3-8B's (paired bootstrap; $^{**}p{<}.01$, $^{***}p{<}.001$).}
\label{tab:edits}
\end{table}

\paragraph{Other query properties.} In a per-query logistic model over all 502 known-item queries, controlling for query length, proper nouns, target-poem length, target confusability (similarity of the target to its nearest neighbour in the corpus) and corpus leaves the containment coefficient essentially unchanged for all ten systems, and the overlap features account for $59$ to $85\%$ of the explained variance (Appendix~\ref{app:robust}).

\begin{table}[t]
\centering\small
\setlength{\tabcolsep}{2.1pt}
\begin{tabular}{l cccc cc}
\toprule
& \multicolumn{2}{c}{FannOrFlop R@5} & \multicolumn{2}{c}{slope} & \multicolumn{2}{c}{$\rho$} \\
\cmidrule(lr){2-3}\cmidrule(lr){4-5}\cmidrule(lr){6-7}
System & quote & no-qt. & $\Delta$ & vs.\,ref & lex. & emb. \\
& \scriptsize $n{=}80$ & \scriptsize $n{=}164$ & & & & \\
\midrule
Qwen3-8B & 0.863 & 0.646 & 0.22 & ref. & 0.17 & \textbf{0.32} \\
BGE-M3 & 0.875 & 0.604 & 0.27 & $+0.06$ & 0.22 & \textbf{0.35} \\
mE5-large & 0.713 & 0.512 & 0.20 & $-0.02$ & 0.25 & \textbf{0.46} \\
jina-v3 & 0.787 & 0.579 & 0.21 & $-0.01$ & 0.29 & \textbf{0.45} \\
silma-sts & 0.750 & 0.476 & 0.27 & $+0.06$ & 0.32 & \textbf{0.36} \\
gte & 0.775 & 0.457 & 0.32 & $+0.10$ & 0.32 & \textbf{0.60}$^{\dagger}$ \\
Qwen3-0.6B & 0.700 & 0.476 & 0.22 & $+0.01$ & 0.30 & \textbf{0.42} \\
LaBSE & 0.525 & 0.207 & 0.32 & $+0.10$ & 0.37 & \textbf{0.50} \\
Muffakir-v2 & 0.300 & 0.165 & 0.14 & $-0.08$ & 0.30 & \textbf{0.36} \\
\midrule
\textbf{BM25} & 1.000 & 0.622 & 0.38 & \textbf{$+0.16$$^{**}$} & \textbf{0.61} & 0.35 \\
\bottomrule
\end{tabular}
\caption{Left: verse-mode Recall@5 on \textit{Fann or Flop}, stratified by verbatim quotation, with slope $\Delta$ and its difference against Qwen3-8B (paired bootstrap). Right: within-query Spearman $\rho$ between overlap change and retrieval change (\textit{Diwan}, AI explanations); bold marks the better predictor. $\dagger$gte's $\rho_{\mathrm{emb}}$ is not independent of the measure (Appendix~\ref{app:pred}).}
\label{tab:controlled}
\end{table}

\subsection{A graded mechanism}
\label{sec:graded}

\begin{figure}[t]
\centering
\includegraphics[width=\columnwidth]{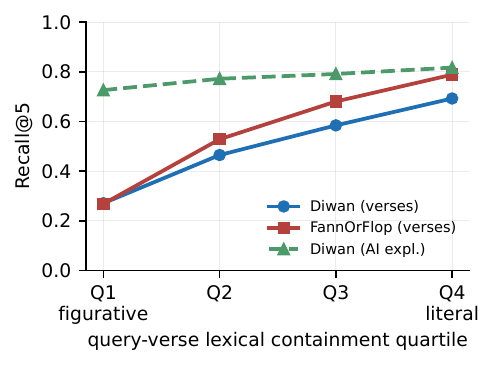}\\[6pt]
\includegraphics[width=\columnwidth]{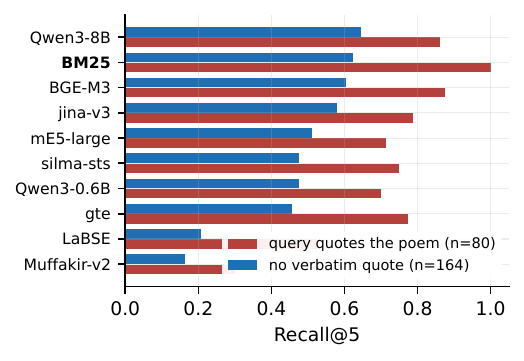}
\caption{Top: Verse-mode Recall@5 rises monotonically with query-verse containment on both corpora; AI explanations flatten the curve on \textit{Diwan}. Bottom: The same systems on \textit{Fann or Flop}, split by whether the query quotes the poem.}
\label{fig:mech}
\end{figure}

The augmentation effect follows the same axis: substituting AI explanations costs BM25 $55$ points on quoting queries but only $31$ without them, and becomes positive for several dense encoders once quotation is removed. The effect is graded, not binary (Figure~\ref{fig:mech}a): verse-mode Recall@5 rises monotonically across query-verse containment quartiles on both corpora, and AI explanations flatten the \textit{Diwan} curve, lifting the most figurative quartile from $0.271$ to $0.726$. Augmentation helps most where the raw verse offers the least.

Retrieval here is not only surface matching. For each query we compute the change in query-document overlap produced by an explanation under two measures: lexical containment and the cosine between query and document embeddings. We then correlate each with the per-query change in retrieval success (Table~\ref{tab:controlled}, right; Figure~\ref{fig:pred}).

Both measures predict retrieval success, and every correlation is positive, but they split cleanly by system type. For BM25, lexical containment is the stronger predictor ($\rho=0.61$ vs.\ $0.35$ on \textit{Diwan}), and its lexical $\rho$ exceeds that of every dense encoder in all three settings. The embedding measure is the better predictor for all nine dense encoders on \textit{Diwan} and eight of nine on \textit{Fann or Flop} (a near-tie for Qwen3-0.6B). BM25's gains track the words an explanation adds in common with the query, while the dense encoders' gains track how much closer it brings the document in embedding space. Word overlap alone would make the effect look purely lexical.

\section{Track B: Thematic Retrieval}
\label{sec:trackb}

Track A measures whether a system can recover a specific poem, but cannot say whether retrieved poems are actually about what was asked. That is what an exploratory user wants, and what Track B evaluates. Here we score each system's top-3 results on the 200 blind thematic queries, for which no gold poem exists, using Gemini 3.5 Flash as judge on seven aspects scored from 1 to 5. The judge sees the first six verses of each retrieved poem. Track B claims are restricted to verse mode, where no model-generated text is shown to the judge, so self-preference toward our AI explanations, which a different version of the same model family generated, cannot arise.

\paragraph{The judge agrees with blind human judgments.} Twenty thematic queries were pooled across
BM25, jina-v3 and their fusion, deduplicated to 126 poems, shuffled with system identity removed,
and graded by two authors independently, who were told that lexical overlap is not evidence of
relevance (Appendix~\ref{app:human}). The raters agree moderately (quadratic-weighted
$\kappa{=}0.53$, $90\%$ within one point). Across the 40 system-query cells covered by both, the
first rater's grades and the judge's overall-quality score agree strongly (Spearman $\rho{=}0.75$,
$p{=}2\times10^{-8}$; $\rho{=}0.64$ for the second rater), and both raters rank the annotated
systems as the judge does. Human raters who ignore surface matching reach the judge's ordering,
which argues against the concern that the rubric alone produces it.

\paragraph{Results.} Table~\ref{tab:trackb} gives Track B scores for all ten systems on both
corpora, complete-case over all 200 queries with no judge failures. The picture is the reverse of
Track A's. \textbf{BM25 is last of ten on both corpora} ($1.90$ on \textit{Fann or Flop}, $1.70$ on
\textit{Diwan}), and is beaten significantly by eight of the nine dense encoders on each (paired
Wilcoxon, Holm-corrected; $\Delta$ up to $-1.04$ against Qwen3-8B, $p{=}2\times10^{-18}$). The
exception on both corpora is LaBSE, the weakest dense encoder, which BM25 is statistically
indistinguishable from ($p{=}.35$ on \textit{Diwan}, $p{=}.054$ on \textit{Fann or Flop}). The blind
human judgments give the same ordering on the systems they cover: on \textit{Fann or Flop}, jina-v3
$0.93$, the RRF hybrid $0.83$, BM25 $0.50$ for the first rater (paired Wilcoxon BM25 vs.\ jina-v3
$p{=}.016$; pooled over both corpora $\Delta{=}{-}0.45$, $p{=}.005$).

\paragraph{Two open-weights judges replicate it.} A proprietary judge cannot be re-run once
withdrawn, so we repeat Track B locally over a random half of the thematic queries ($2{,}000$ cells
per judge). The second judge, Gemma-3-27B-it, uses the identical rubric. It agrees closely with the primary
judge at the cell level (Spearman $\rho{=}0.67$ on overall quality) and on system ordering (Kendall
$\tau{=}{+}0.85$ on \textit{Diwan}, ${+}0.81$ on \textit{Fann or Flop}), and it tracks the blind
human judgments too ($\rho{=}0.70$). The main difference is at the bottom: BM25 and LaBSE swap
places, so BM25 is \emph{ninth} rather than last, still significantly below the other eight dense
encoders on both corpora. The primary judge could not separate these two systems either. Because
both judges are Google models, we add a third from another family, Qwen3-32B. It also places BM25
ninth of ten on both corpora, above LaBSE only ($\tau{=}0.67$ and $0.89$ with the primary judge).
We then rerun it with a neutral rubric, which drops the one aspect that treats literal matching as a
weakness, together with the sentence asking the judge to reward implicit meaning; the rest of the
prompt is unchanged. Under this neutral rubric BM25 is tenth on \textit{Fann or Flop} and ninth on \textit{Diwan}, and the ordering
barely moves ($\tau{=}0.90$ and $0.97$ against the original rubric). Dropping the aspect raises
every system's score, BM25's the least ($+0.10$ against up to $+0.27$). The literal-match aspect is
therefore not what places BM25 near the bottom (Appendix~\ref{app:judge2}).

\paragraph{The judge separates matching from aboutness.} BM25 is last of ten on almost every aspect
requiring knowledge of what a poem is about, with its largest deficits on metaphor understanding and
semantic coverage; but on \emph{direct textual relevance}, which measures how far retrieved poems
merely repeat the query's wording, it is \emph{first} of ten on both corpora ($4.26$, $4.48$;
second of ten under the second judge, behind LaBSE; Appendix~\ref{app:aspects}). The judge
identifies BM25 as the most surface-matching system while ranking it last overall. This supports
\S\ref{sec:bridge} through a channel that shares nothing with our lexical statistics, and it argues
against the reading that the judge merely penalises lexical retrieval.

\begin{table}[t]
\centering\small
\setlength{\tabcolsep}{3.5pt}
\begin{tabular}{l cc c cc}
\toprule
& \multicolumn{2}{c}{\textit{FannOrFlop}} & & \multicolumn{2}{c}{\textit{Diwan}} \\
\cmidrule(lr){2-3}\cmidrule(lr){5-6}
System & Track A & Track B & & Track A & Track B \\
\midrule
Qwen3-8B & 0.717 & 2.94 & & 0.748 & 2.83 \\
BGE-M3 & 0.693 & 2.47 & & 0.574 & 2.29 \\
mE5-large & 0.578 & 2.52 & & 0.585 & 2.35 \\
jina-v3 & 0.647 & 2.64 & & 0.523 & 2.53 \\
silma-sts & 0.566 & 2.75 & & 0.558 & 2.63 \\
gte & 0.561 & 2.37 & & 0.488 & 2.18 \\
Qwen3-0.6B & 0.549 & 2.35 & & 0.527 & 2.14 \\
LaBSE & 0.311 & 2.07 & & 0.368 & 1.75 \\
Muffakir-v2 & 0.209 & 2.43 & & 0.233 & 2.27 \\
\midrule
\textbf{BM25} & 0.746 & 1.90 & & 0.392 & 1.70 \\
\midrule
\multicolumn{6}{l}{\textit{blind human relevance of top-3 (0 to 2), 20 queries}} \\
jina-v3 & \multicolumn{2}{c}{0.93} & & \multicolumn{2}{c}{0.87} \\
hybrid & \multicolumn{2}{c}{0.83} & & \multicolumn{2}{c}{0.57} \\
\textbf{BM25} & \multicolumn{2}{c}{0.50} & & \multicolumn{2}{c}{0.40} \\
\bottomrule
\end{tabular}
\caption{Track A (known-item Recall@5) beside Track B (thematic judge score, 1 to 5), verse mode,
$n{=}200$ queries per cell with no judge failures. The lower block gives the first rater's blind
human relevance of the top-3 (0 to 2) for the three annotated systems.}
\label{tab:trackb}
\end{table}

\section{The Dissociation}
\label{sec:dissoc}

\begin{figure}[t]
\centering
\includegraphics[width=\columnwidth]{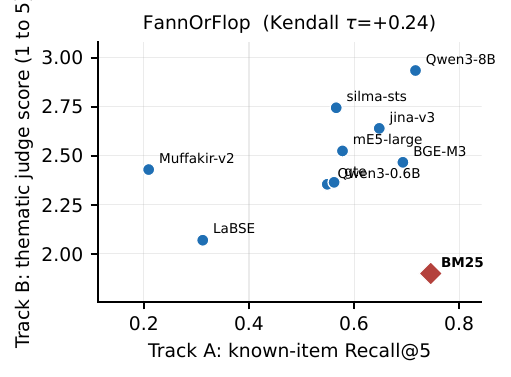}\\[6pt]
\includegraphics[width=\columnwidth]{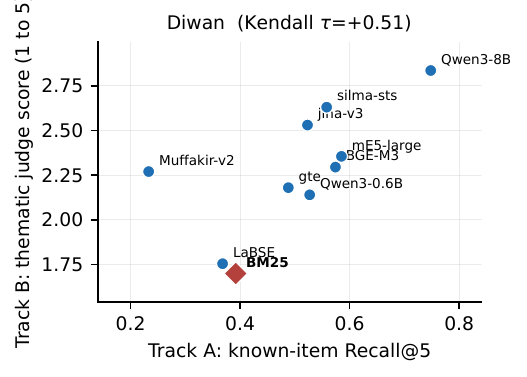}
\caption{The dissociation. Each point is a system; $x$ is known-item Recall@5, $y$ is the thematic
judge score. Top: on \textit{Fann or Flop}, BM25 (red diamond) is the best system on one axis and
the worst on the other. Bottom: on \textit{Diwan}, where no query carries remembered wording, the
two protocols broadly agree.}
\label{fig:dissoc}
\end{figure}

Tracks A and B measure the same ten systems, in the same indexing mode, over the same corpora. They do not agree, and the disagreement falls where the mechanism predicts. On \textit{Fann or Flop}, BM25 ranks first of ten on Track A and last of ten on Track B, a shift of nine places (Figure~\ref{fig:dissoc}). No other system moves more than four. On \textit{Diwan}, where queries carry no remembered wording, BM25 is poor under both protocols and the ranks broadly agree. The dissociation is a property of the tip-of-the-tongue regime, not of BM25 in general.

Correlating the two rankings gives Kendall $\tau{=}+0.24$ ($p{=}0.38$) on \textit{Fann or Flop}
against $\tau{=}+0.51$ ($p{=}0.047$) on \textit{Diwan}. We do not read the first as evidence that
the protocols are unrelated in general: with ten systems one outlier is enough to move $\tau$, and
removing BM25 restores agreement ($\tau{=}+0.56$, $p{=}0.045$). Among dense encoders the protocols
largely concur; what dissociates is the lexical system, and only where remembered wording lets
surface matching succeed without any notion of aboutness.

The two results are consistent. Track A rewards recovering a specific remembered item, and where the query carries the item's wording, surface matching is close to sufficient. Track B asks whether retrieved poems are about a theme that is never stated in the poems' own words, and there surface matching has nothing to work with. Both protocols are valid, but they measure different things, and a system's score under one carries little information about its score under the other.

Reporting a single protocol can therefore mislead. A study that evaluated only known-item retrieval on a tip-of-the-tongue query set would conclude that dense retrieval is not worth its cost for Arabic poetry, since a bag of words beats an 8B multilingual encoder. A study that evaluated only thematic relevance would conclude the opposite, with BM25 last of ten and significantly behind eight of the nine dense encoders. Each conclusion is supported by its own data, and neither holds in general.

\section{Fusion and Deployment}
\label{sec:recipe}

If the query register decides which signal is useful, a practical system should combine signals rather than choose one. We test reciprocal-rank fusion (RRF; \citealp{cormack2009rrf}), which merges several ranked lists into one, combining a verse index, an explanation index and BM25 (Table~\ref{tab:fusion}, strongest encoder). Fusing explanations with verses helps everywhere, making it a safe default. Adding BM25's list does not. Comparing the best configuration that includes BM25 against the best that excludes it, BM25 is worth $+6.1$ points on \textit{Fann or Flop} ($0.885$ vs.\ $0.824$) and costs $9.7$ on \textit{Diwan} ($0.833$ vs.\ $0.930$, where the explanation index alone is best). The best \textit{Fann or Flop} configuration reaches $0.885$, against $0.816$ for the best single index. The pattern holds with every dense encoder in place of Qwen3-8B: BM25 helps on \textit{Fann or Flop} for all nine ($+6.1$ to $+23.4$ points) and hurts on \textit{Diwan} for all nine ($-2.7$ to $-9.7$). The gain is not limited to quoting queries: on the non-quoting \textit{Fann or Flop} queries, adding BM25 still lifts Qwen3-8B from $0.646$ to $0.756$, while on \textit{Diwan}, where BM25 alone is weak, it lowers Qwen3-8B from $0.748$ to $0.616$.

No single recipe is best everywhere. An archive serving readers who half-remember lines should include BM25; one serving thematic browsing should not.

\begin{table}[t]
\centering\small
\setlength{\tabcolsep}{3.5pt}
\begin{tabular}{l ccc c c}
\toprule
& \multicolumn{3}{c}{\textit{FannOrFlop}} & & \textit{Diwan} \\
\cmidrule(lr){2-4}\cmidrule(lr){6-6}
Index configuration & all & quote & no-qt. & & all \\
\midrule
verses only & 0.717 & 0.863 & 0.646 & & 0.748 \\
AI expl.\ only & 0.656 & 0.713 & 0.628 & & 0.930 \\
FoF expl.\ only & 0.816 & 0.812 & 0.817 & & n/a \\
vrs.\,$\oplus$\,AI expl. & 0.730 & 0.800 & 0.695 & & 0.857 \\
vrs.\,$\oplus$\,FoF expl. & 0.824 & 0.912 & 0.780 & & n/a \\
vrs.\,$\oplus$\,BM25 & 0.836 & 1.000 & 0.756 & & 0.616 \\
vrs.\,$\oplus$\,BM25\,$\oplus$\,AI expl. & 0.873 & 1.000 & 0.811 & & 0.833 \\
vrs.\,$\oplus$\,BM25\,$\oplus$\,FoF expl. & 0.885 & 0.988 & 0.835 & & n/a \\
\bottomrule
\end{tabular}
\caption{Reciprocal-rank fusion over index configurations, Qwen3-8B. $\oplus$ denotes RRF ($k{=}60$). Fusing explanations is safe everywhere; adding BM25 helps only where queries carry remembered wording.}
\label{tab:fusion}
\end{table}

\section{Discussion}
\label{sec:discussion}

A single Recall@5 figure for ``semantic search over Arabic poetry'' is a joint property of a system and the register of the queries it was measured on, and the two standard protocols disagree on the same systems.

On the benchmarking side, query provenance should be reported and, ideally, manipulated. The verbatim-overlap statistic we use costs nothing to compute and exposes the confound directly; we recommend reporting it alongside any known-item result on literary text. Known-item retrieval and thematic relevance should also be reported together, since each alone supports a defensible but opposite claim about whether dense retrieval is worth its cost.

Our queries were written by the authors. The results therefore show how retrieval systems behave under quotation-like and abstract queries, not how often each kind occurs among real users of a poetry archive. Measuring that mix with real users is the natural next step, and would tell an archive which recipe of \S\ref{sec:recipe} to deploy.

The queries are also in contemporary Arabic, while the poems range from the pre-Islamic period to the modern era. On \textit{Fann or Flop}, at equal lexical overlap and query length, dense encoders do worse on older poetry ($+0.27$ log-odds of success per era step towards the present, $p{=}.010$), while BM25 shows no such trend ($p{=}.14$; interaction $p{=}.015$; Appendix~\ref{app:corpus}). This fits a temporal gap between modern queries and classical vocabulary that affects learned representations more than exact matching. It rests on one corpus and four coarse era bins, so we treat it as exploratory.

On the modelling side, the open problem is the non-quoting, most figurative queries: on verses the best system reaches about $0.65$ there ($0.82$ with FoF explanations, which only one corpus has).

\section{Conclusion}

We presented what is, to our knowledge, the first systematic study of retrieval over Classical Arabic poetry. Which retriever wins, and whether explanation augmentation helps, is governed by how much of a poem's wording the query still carries. Editing the same queries confirms it: added wording helps the lexical retriever far more than the strongest dense encoder, and removed wording erases its lead. A corpus-level ranking therefore reports its query mix as much as its systems. The two evaluation protocols dissociate, as two human annotators and three LLM judges confirm.

\section*{Limitations}

Known-item evaluation assumes one correct poem per query. A blind spot check finds that about a third of Qwen3-8B's misses are acceptable answers, more often on abstract queries, so absolute Recall is conservative; paired comparisons are unaffected and the quotation interaction keeps its sign (Appendix~\ref{app:human}). Our queries were written by the authors rather than collected from real users, so the endpoints of the spectrum are constructed and we cannot say how often each register occurs in practice; the hand rewrites also come from a single co-author. Human relevance annotation covers 126 items with moderate agreement between two raters ($\kappa{=}0.53$), enough for the overall Track~B ordering but not for fine-grained comparisons, and the open-weights judges cover half of the thematic queries. The AI explanations come from one model, since withdrawn from its API, and $8\%$ of those audited misread the verse, although BM25 shows the same augmentation pattern regardless of explanation quality. The \textit{Diwan} sample over-represents frequent poets and contains duplicated poems, which change no reported number by more than $0.4$ points (Appendix~\ref{app:corpus}).

\section*{Ethical Considerations}

This work supports the preservation, exploration and teaching of Classical Arabic poetry. Automated retrieval and model-generated explanations can misread metaphorical, religious or culturally sensitive language; our own analysis shows that current systems often match on surface form rather than meaning, and that a minority of generated explanations misread the verse, so such tools should aid rather than replace human literary expertise. Our \textit{Diwan} sample over-represents frequently anthologised poets and may underperform on marginal traditions. Annotation was performed by two of the authors; no external annotators were recruited or compensated.

\section*{AI Usage Disclosure}

AI assistants were used for coding support and for drafting and editing prose. All experiments were designed and run by the authors, and every number in this paper is recomputed from per-query logs by the released scripts.

\section*{Acknowledgments}

We thank Sukai Labs for providing access to the NVIDIA GeForce RTX 5090 GPU on which our local experiments were run.

\bibliography{refs}

\appendix

\section{Reproducibility}
\label{app:repro}
Code, per-query ranked lists, all judges' per-query outputs, the annotation and spot-check guidelines with their completed judgments, the edited queries, and the query benchmark are available at \url{https://github.com/Ellzo/what-the-query-remembers}. Every number is recomputed from per-query rank dumps by released scripts. The pipeline emits one file per (system, corpus, indexing mode, query set) containing the top-100 poem ids, scores and the rank of the gold poem; all metrics, confidence intervals, significance tests, stratifications, fusion variants and figures are derived from those files.

\paragraph{Computational budget.} All local computation ran on one NVIDIA GeForce RTX 5090 (32\,GB). No model was trained or fine-tuned; every model is used off the shelf for inference. Indexing and retrieval for the nine dense encoders, over both corpora, all indexing modes and both query sets, takes $1{,}568$\,s of GPU time in total, of which Qwen3-Embedding-8B alone accounts for $852$\,s; BM25 runs on CPU in under a minute. The second Track~B judge, Gemma-3-27B-it quantised to 4 bits, takes $66$ minutes for its $2{,}000$ judgments at batch size 6. The third judge, Qwen3-32B quantised to 4 bits, runs on the same GPU. The normalisation ablation of Appendix~\ref{app:norm} adds under $40$ minutes of wall clock, including model downloads. Figures reported here are wall clock on an otherwise idle machine.

\section{Normalisation Ablation}
\label{app:norm}
Measured on \textit{Diwan}, verse mode, under three settings: raw text with diacritics retained; light normalisation, stripping diacritics and tatweel; and full normalisation, which additionally unifies \textit{alef}/\textit{ya}/\textit{ta-marbuta} and deletes non-Arabic characters.

\begin{center}\small
\begin{tabular}{lccc}
\toprule
Recall@5 & raw & light & full \\
\midrule
jina-v3 & 0.178 & \textbf{0.523} & 0.434 \\
gte & 0.023 & \textbf{0.488} & 0.469 \\
LaBSE & 0.093 & \textbf{0.368} & 0.333 \\
\bottomrule
\end{tabular}
\end{center}

Diacritic removal is essential, and the effect is very large and consistent across all three encoders ($+27$ to $+47$ Recall@5 points), which is unsurprising: \textit{Diwan} verses are diacritised and the models' subword vocabularies are not. The further orthographic unification points the same way for all three but varies in size by a factor of four, costing $8.9$ points on jina-v3, $3.5$ on LaBSE and $1.9$ on gte. It is also not uniformly harmful: gte's Recall@1 improves under it ($0.256\!\rightarrow\!0.318$) while LaBSE's falls ($0.209\!\rightarrow\!0.167$). We therefore apply light normalisation to every dense encoder, which is the better Recall@5 setting for all three, and full normalisation to BM25 and the lexical overlap measures, where collapsing orthographic variants is standard. The nine-point cost measured on jina-v3 should not be generalised to dense encoders as a class.

\section{Which Overlap Measure Predicts}
\label{app:pred}
\begin{figure}[h]
\centering
\includegraphics[width=\columnwidth]{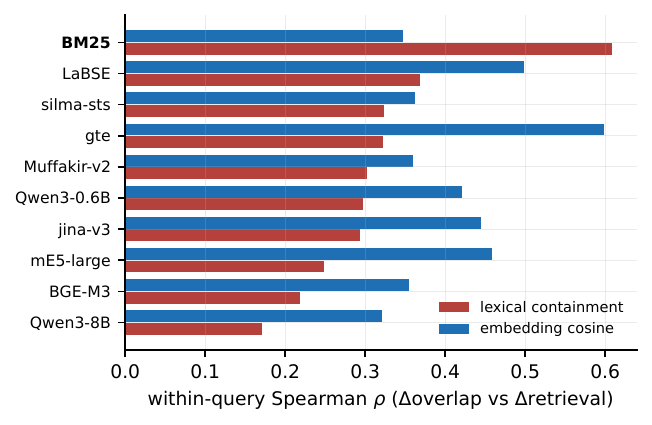}
\caption{Within-query $\rho$ per system under a lexical and an embedding overlap measure.}
\label{fig:pred}
\end{figure}

Figure~\ref{fig:pred} shows, per system, the within-query Spearman correlation between the change in query-document overlap induced by an explanation and the change in retrieval success, under a lexical and an embedding measure (\textit{Diwan}, AI-explanation mode). BM25 is the only system for which the lexical measure is the stronger predictor.

Among dense encoders on \textit{Diwan}, the more capable the model, the less its behaviour is governed by surface overlap ($\rho=-0.73$ between verse-mode Recall@5 and lexical $\rho$, $p=.025$, $n=9$). This does not hold on \textit{Fann or Flop} ($p>0.5$), so it is a property of the abstract-query regime only, and it rests on nine points.

One caveat on the embedding measure. It is computed with a single fixed encoder (gte) for every system, so that all systems are scored on the same axis. gte is itself one of the ten systems, and its own $\rho_{\mathrm{emb}}$ is consequently the highest in all three settings ($0.60$, $0.56$, $0.57$ against a next-highest of $0.50$, $0.41$, $0.38$); that row is not independent of the measure and should not be read as evidence about gte. Every conclusion above survives dropping it: the embedding measure still predicts better for the remaining eight dense encoders on \textit{Diwan} and seven of eight on \textit{Fann or Flop}, and BM25 remains the only system whose behaviour the lexical measure predicts best, with a lexical $\rho$ above every dense encoder's.

\section{Multiple Comparisons}
\label{app:mult}
We report $57$ hypothesis tests in six families: the mode contrasts of Table~\ref{tab:main} (one family per corpus and contrast, ten systems each), the quotation-slope differences of \S\ref{sec:strat} (nine systems against the reference encoder), and the Track~B comparisons of BM25 against each dense encoder (one family per corpus). Holm-Bonferroni correction within each family leaves $36$ of the $39$ nominally significant tests significant. The three that do not survive are all \textit{Fann or Flop} AI-explanation contrasts, for BGE-M3 ($p{=}.009\!\rightarrow\!.051$), mE5-large ($.033\!\rightarrow\!.136$) and Muffakir-v2 ($.027\!\rightarrow\!.136$); Table~\ref{tab:main} shows them unstarred. No claim in the paper depends on them. The interaction of \S\ref{sec:strat} is the smallest $p$-value in its family, so Holm and Bonferroni coincide there at $p{=}.045$. Per-test and corrected values for all $57$ are released as \texttt{eacl\_multiplicity.csv}. The edited-query tests of \S\ref{sec:edits} and the analyses of Appendices~\ref{app:robust} and~\ref{app:corpus} are reported uncorrected, except the metric-robustness test, which is Holm-corrected across the nine comparisons within each metric.

\section{Threshold Robustness of the Stratification}
\label{app:threshold}
\S\ref{sec:strat} splits \textit{Fann or Flop} on whether the query shares a verbatim three-token span with the target poem. Table~\ref{tab:appthreshold} repeats the split at spans of two to five tokens. Two results hold at every threshold: BM25 retrieves the gold poem within the top five for \emph{every} quoting query at the three-, four- and five-token thresholds (and for $140$ of $146$ at two tokens), and its quotation slope stays significantly steeper than Qwen3-8B's ($p{=}.0001$, $.005$, $.006$, $.049$ for thresholds two to five). The non-quoting comparison behaves as the graded account predicts: at a two-token threshold the non-quoting subset is the most abstract and Qwen3-8B leads it significantly ($0.571$ vs.\ $0.429$, $p{=}.039$); as the threshold rises, queries with two- and three-token overlaps re-enter the non-quoting subset and the gap closes to parity. BM25's slope is bounded above by its perfect quoting-subset score, so at the four- and five-token thresholds, where its non-quoting score has risen to $0.70$ and $0.72$, three weaker encoders show nominally larger slopes; its slope remains significantly larger than that of the strongest dense encoder throughout.

\begin{table}[h]
\centering\scriptsize
\setlength{\tabcolsep}{2.5pt}
\begin{tabular}{lcccccccc}
\toprule
& \multicolumn{2}{c}{$\geq2$} & \multicolumn{2}{c}{$\geq3$} & \multicolumn{2}{c}{$\geq4$} & \multicolumn{2}{c}{$\geq5$} \\
\cmidrule(lr){2-3}\cmidrule(lr){4-5}\cmidrule(lr){6-7}\cmidrule(lr){8-9}
System & qt. & no-qt. & qt. & no-qt. & qt. & no-qt. & qt. & no-qt. \\
& \scriptsize 146 & \scriptsize 98 & \scriptsize 80 & \scriptsize 164 & \scriptsize 40 & \scriptsize 204 & \scriptsize 24 & \scriptsize 220 \\
\midrule
Qwen3-8B & 0.815 & 0.571 & 0.863 & 0.646 & 0.825 & 0.696 & 0.833 & 0.705 \\
BGE-M3 & 0.856 & 0.449 & 0.875 & 0.604 & 0.950 & 0.642 & 0.958 & 0.664 \\
mE5-large & 0.678 & 0.429 & 0.713 & 0.512 & 0.775 & 0.539 & 0.917 & 0.541 \\
jina-v3 & 0.801 & 0.418 & 0.787 & 0.579 & 0.850 & 0.608 & 0.917 & 0.618 \\
silma-sts & 0.692 & 0.378 & 0.750 & 0.476 & 0.725 & 0.534 & 0.792 & 0.541 \\
gte & 0.753 & 0.276 & 0.775 & 0.457 & 0.850 & 0.505 & 0.917 & 0.523 \\
Qwen3-0.6B & 0.678 & 0.357 & 0.700 & 0.476 & 0.775 & 0.505 & 0.833 & 0.518 \\
LaBSE & 0.445 & 0.112 & 0.525 & 0.207 & 0.600 & 0.255 & 0.667 & 0.273 \\
Muffakir-v2 & 0.281 & 0.102 & 0.300 & 0.165 & 0.275 & 0.196 & 0.292 & 0.200 \\
\midrule
\textbf{BM25} & 0.959 & 0.429 & 1.000 & 0.622 & 1.000 & 0.696 & 1.000 & 0.718 \\
\bottomrule
\end{tabular}

\caption{Verse-mode Recall@5 on \textit{Fann or Flop}, split at verbatim spans of $\geq$2 to $\geq$5 tokens. Subset sizes are given under each threshold.}
\label{tab:appthreshold}
\end{table}

\section{Construction of the Edited Queries}
\label{app:edits}
All edits apply to \textit{Fann or Flop}, verse mode, and are run through the same pipeline as the main results. Before any edited run, the unedited queries are re-run and must reproduce the released ranks of all ten systems exactly.

\paragraph{Adding a half-verse.} Each of the 164 non-quoting queries is extended with the phrase \textarabic{ومن أبياتها:} (``and among its verses:'') followed by one verbatim half-verse of the target poem. Three variants take the half-verse from the opening verse, a middle verse or the final verse. Nothing else in the query changes.

\paragraph{Hand rewrites.} A co-author rewrote the quoted words of each of the 80 quoting queries in their own words, keeping the information need and any proper names, with a live check that no verbatim three-token span of the target poem remained. One query is a quotation from start to end and was marked impossible, leaving 79.

\paragraph{LLM rewrites.} Qwen3-32B rewrote only the quoted span of each quoting query under the same constraints. A second co-author rated every rewrite for whether it kept the information need and whether remembered wording was left. On the $55$ rewrites judged to keep the information need at least in part and to leave at most part of the quoted wording, BM25 loses $0.473$ and Qwen3-8B $0.255$ ($p{=}.006$).

\paragraph{Deletion (control).} Deleting the tokens a quoting query shares with its poem removes information as well as wording and shortens the queries from $10.9$ to $6.4$ tokens on average, so we report it only as a control: BM25 loses $0.600$ and Qwen3-8B $0.413$ ($p{=}.008$).

\section{Metric Robustness and Query Covariates}
\label{app:robust}
Table~\ref{tab:appmetrics} repeats the quotation-slope test of \S\ref{sec:strat} on four metrics. BM25's slope is the steepest of the ten on all four. After Holm correction across the nine comparisons, Recall@5 and MRR are at or below $.05$ and Recall@1 and nDCG@10 are at about $.08$.

\begin{table}[h]
\centering\scriptsize
\setlength{\tabcolsep}{2.5pt}
\begin{tabular}{l cc c c cc}
\toprule
& \multicolumn{2}{c}{slope $\Delta$} & & & & \\
\cmidrule(lr){2-3}
Metric & BM25 & Qwen3-8B & difference & 95\% CI & $p$ & $p_{\mathrm{Holm}}$ \\
\midrule
R@1 & 0.419 & 0.237 & $+0.182$ & [0.040, 0.325] & 0.009 & 0.085 \\
R@5 & 0.378 & 0.216 & $+0.162$ & [0.046, 0.285] & 0.005 & 0.045 \\
MRR & 0.389 & 0.229 & $+0.160$ & [0.045, 0.277] & 0.006 & 0.050 \\
nDCG@10 & 0.371 & 0.231 & $+0.140$ & [0.035, 0.247] & 0.009 & 0.079 \\
\bottomrule
\end{tabular}

\caption{Quotation slope (quoting minus non-quoting subset) on \textit{Fann or Flop}, verse mode, and BM25's difference against Qwen3-8B (paired bootstrap, $10^4$ resamples).}
\label{tab:appmetrics}
\end{table}

For the covariate analysis we fit, per system, a logistic model of Recall@5 success over the 502 known-item queries, with z-scored predictors and standard errors clustered by query. The controls are query length, the number of proper nouns in the query (tagged by Qwen3-32B, a noisy covariate), target-poem length, target confusability (TF-IDF cosine between the target and its nearest other poem, computed without the query) and a corpus indicator. Adding them changes the containment coefficient by less than $0.07$ log-odds per standard deviation for every system, and it stays significant at $p<10^{-4}$ for all ten. A Shapley decomposition attributes $59$ to $85\%$ of each system's explained variance to the overlap features (containment, longest shared span, quoted fraction). In a pooled model, the interaction of containment with BM25 is $+2.99$ log-odds per standard deviation ($p<10^{-15}$).

\section{Full Track A Results}
\label{app:full}
Table~\ref{tab:apptracka} gives Recall@1/@5, MRR and nDCG@10 with bootstrap 95\% confidence intervals ($10^4$ resamples) for every system, corpus and indexing mode. Per-subset results are released with the code as \texttt{eacl\_trackA.csv}.

\begin{table*}[t]
\centering\scriptsize
\setlength{\tabcolsep}{3pt}
\begin{tabular}{ll cccc}
\toprule
System & Mode & R@1 & R@5 & MRR & nDCG@10 \\
\midrule
\multicolumn{6}{l}{\textit{Fann or Flop}} \\
Qwen3-8B & verses & 0.566 \scriptsize [0.504,0.627] & 0.717 \scriptsize [0.660,0.770] & 0.635 \scriptsize [0.581,0.688] & 0.667 \scriptsize [0.615,0.718] \\
Qwen3-8B & AI expl. & 0.426 \scriptsize [0.365,0.488] & 0.656 \scriptsize [0.594,0.713] & 0.530 \scriptsize [0.475,0.584] & 0.570 \scriptsize [0.517,0.622] \\
Qwen3-8B & FoF expl. & 0.639 \scriptsize [0.578,0.701] & 0.816 \scriptsize [0.766,0.865] & 0.725 \scriptsize [0.677,0.773] & 0.754 \scriptsize [0.710,0.800] \\
BGE-M3 & verses & 0.574 \scriptsize [0.512,0.635] & 0.693 \scriptsize [0.635,0.750] & 0.634 \scriptsize [0.578,0.689] & 0.657 \scriptsize [0.603,0.712] \\
BGE-M3 & AI expl. & 0.406 \scriptsize [0.344,0.467] & 0.598 \scriptsize [0.537,0.660] & 0.489 \scriptsize [0.434,0.544] & 0.515 \scriptsize [0.459,0.570] \\
BGE-M3 & FoF expl. & 0.566 \scriptsize [0.504,0.627] & 0.725 \scriptsize [0.668,0.779] & 0.646 \scriptsize [0.593,0.699] & 0.676 \scriptsize [0.624,0.727] \\
mE5-large & verses & 0.414 \scriptsize [0.352,0.475] & 0.578 \scriptsize [0.516,0.639] & 0.485 \scriptsize [0.429,0.542] & 0.510 \scriptsize [0.454,0.566] \\
mE5-large & AI expl. & 0.332 \scriptsize [0.275,0.389] & 0.492 \scriptsize [0.430,0.553] & 0.408 \scriptsize [0.355,0.464] & 0.437 \scriptsize [0.382,0.492] \\
mE5-large & FoF expl. & 0.557 \scriptsize [0.492,0.619] & 0.717 \scriptsize [0.660,0.770] & 0.632 \scriptsize [0.576,0.685] & 0.662 \scriptsize [0.611,0.714] \\
jina-v3 & verses & 0.479 \scriptsize [0.418,0.541] & 0.647 \scriptsize [0.586,0.705] & 0.559 \scriptsize [0.502,0.614] & 0.589 \scriptsize [0.533,0.643] \\
jina-v3 & AI expl. & 0.303 \scriptsize [0.246,0.361] & 0.504 \scriptsize [0.439,0.566] & 0.395 \scriptsize [0.343,0.449] & 0.430 \scriptsize [0.377,0.486] \\
jina-v3 & FoF expl. & 0.512 \scriptsize [0.451,0.574] & 0.676 \scriptsize [0.615,0.734] & 0.592 \scriptsize [0.538,0.645] & 0.623 \scriptsize [0.569,0.676] \\
silma-sts & verses & 0.389 \scriptsize [0.332,0.451] & 0.566 \scriptsize [0.504,0.627] & 0.479 \scriptsize [0.424,0.532] & 0.511 \scriptsize [0.455,0.566] \\
silma-sts & AI expl. & 0.361 \scriptsize [0.299,0.422] & 0.508 \scriptsize [0.447,0.570] & 0.439 \scriptsize [0.384,0.495] & 0.475 \scriptsize [0.419,0.529] \\
silma-sts & FoF expl. & 0.516 \scriptsize [0.451,0.578] & 0.717 \scriptsize [0.660,0.775] & 0.606 \scriptsize [0.552,0.659] & 0.641 \scriptsize [0.590,0.694] \\
gte & verses & 0.439 \scriptsize [0.377,0.500] & 0.561 \scriptsize [0.500,0.623] & 0.504 \scriptsize [0.448,0.561] & 0.532 \scriptsize [0.477,0.589] \\
gte & AI expl. & 0.254 \scriptsize [0.201,0.307] & 0.389 \scriptsize [0.328,0.451] & 0.331 \scriptsize [0.280,0.382] & 0.356 \scriptsize [0.304,0.409] \\
gte & FoF expl. & 0.459 \scriptsize [0.398,0.520] & 0.635 \scriptsize [0.574,0.693] & 0.541 \scriptsize [0.485,0.596] & 0.569 \scriptsize [0.513,0.624] \\
Qwen3-0.6B & verses & 0.414 \scriptsize [0.352,0.475] & 0.549 \scriptsize [0.488,0.611] & 0.484 \scriptsize [0.427,0.542] & 0.509 \scriptsize [0.452,0.566] \\
Qwen3-0.6B & AI expl. & 0.266 \scriptsize [0.213,0.324] & 0.430 \scriptsize [0.369,0.492] & 0.346 \scriptsize [0.294,0.398] & 0.376 \scriptsize [0.323,0.429] \\
Qwen3-0.6B & FoF expl. & 0.455 \scriptsize [0.393,0.516] & 0.660 \scriptsize [0.598,0.721] & 0.552 \scriptsize [0.498,0.608] & 0.590 \scriptsize [0.537,0.643] \\
LaBSE & verses & 0.197 \scriptsize [0.147,0.250] & 0.311 \scriptsize [0.254,0.373] & 0.253 \scriptsize [0.205,0.303] & 0.272 \scriptsize [0.223,0.322] \\
LaBSE & AI expl. & 0.189 \scriptsize [0.139,0.238] & 0.283 \scriptsize [0.225,0.340] & 0.237 \scriptsize [0.189,0.286] & 0.248 \scriptsize [0.199,0.298] \\
LaBSE & FoF expl. & 0.213 \scriptsize [0.164,0.266] & 0.336 \scriptsize [0.275,0.394] & 0.274 \scriptsize [0.224,0.324] & 0.298 \scriptsize [0.247,0.350] \\
Muffakir-v2 & verses & 0.107 \scriptsize [0.070,0.147] & 0.209 \scriptsize [0.160,0.262] & 0.167 \scriptsize [0.128,0.208] & 0.185 \scriptsize [0.145,0.227] \\
Muffakir-v2 & AI expl. & 0.172 \scriptsize [0.127,0.221] & 0.287 \scriptsize [0.234,0.344] & 0.236 \scriptsize [0.191,0.282] & 0.260 \scriptsize [0.213,0.309] \\
Muffakir-v2 & FoF expl. & 0.324 \scriptsize [0.266,0.385] & 0.467 \scriptsize [0.406,0.529] & 0.399 \scriptsize [0.345,0.455] & 0.427 \scriptsize [0.374,0.483] \\
\textbf{BM25} & verses & 0.656 \scriptsize [0.594,0.717] & 0.746 \scriptsize [0.689,0.799] & 0.699 \scriptsize [0.644,0.751] & 0.720 \scriptsize [0.667,0.770] \\
\textbf{BM25} & AI expl. & 0.246 \scriptsize [0.193,0.303] & 0.357 \scriptsize [0.299,0.418] & 0.301 \scriptsize [0.249,0.354] & 0.323 \scriptsize [0.271,0.377] \\
\textbf{BM25} & FoF expl. & 0.603 \scriptsize [0.541,0.664] & 0.742 \scriptsize [0.684,0.795] & 0.665 \scriptsize [0.611,0.718] & 0.690 \scriptsize [0.635,0.740] \\
\midrule
\multicolumn{6}{l}{\textit{Diwan}} \\
Qwen3-8B & verses & 0.527 \scriptsize [0.469,0.585] & 0.748 \scriptsize [0.694,0.798] & 0.629 \scriptsize [0.580,0.681] & 0.667 \scriptsize [0.618,0.713] \\
Qwen3-8B & AI expl. & 0.779 \scriptsize [0.729,0.830] & 0.930 \scriptsize [0.895,0.961] & 0.849 \scriptsize [0.812,0.885] & 0.871 \scriptsize [0.837,0.903] \\
BGE-M3 & verses & 0.403 \scriptsize [0.341,0.461] & 0.574 \scriptsize [0.512,0.632] & 0.488 \scriptsize [0.436,0.542] & 0.521 \scriptsize [0.467,0.576] \\
BGE-M3 & AI expl. & 0.686 \scriptsize [0.632,0.740] & 0.845 \scriptsize [0.798,0.888] & 0.752 \scriptsize [0.705,0.797] & 0.779 \scriptsize [0.735,0.822] \\
mE5-large & verses & 0.353 \scriptsize [0.298,0.411] & 0.585 \scriptsize [0.523,0.647] & 0.453 \scriptsize [0.401,0.506] & 0.492 \scriptsize [0.440,0.544] \\
mE5-large & AI expl. & 0.659 \scriptsize [0.601,0.713] & 0.802 \scriptsize [0.752,0.849] & 0.726 \scriptsize [0.676,0.774] & 0.753 \scriptsize [0.706,0.798] \\
jina-v3 & verses & 0.345 \scriptsize [0.287,0.403] & 0.523 \scriptsize [0.461,0.581] & 0.429 \scriptsize [0.378,0.482] & 0.456 \scriptsize [0.402,0.509] \\
jina-v3 & AI expl. & 0.589 \scriptsize [0.531,0.647] & 0.756 \scriptsize [0.702,0.806] & 0.667 \scriptsize [0.616,0.717] & 0.694 \scriptsize [0.645,0.742] \\
silma-sts & verses & 0.380 \scriptsize [0.322,0.438] & 0.558 \scriptsize [0.500,0.616] & 0.466 \scriptsize [0.412,0.519] & 0.500 \scriptsize [0.449,0.552] \\
silma-sts & AI expl. & 0.659 \scriptsize [0.601,0.717] & 0.810 \scriptsize [0.764,0.857] & 0.726 \scriptsize [0.677,0.775] & 0.751 \scriptsize [0.705,0.796] \\
gte & verses & 0.256 \scriptsize [0.205,0.310] & 0.488 \scriptsize [0.430,0.550] & 0.361 \scriptsize [0.313,0.411] & 0.402 \scriptsize [0.352,0.452] \\
gte & AI expl. & 0.651 \scriptsize [0.593,0.709] & 0.806 \scriptsize [0.760,0.853] & 0.724 \scriptsize [0.675,0.772] & 0.747 \scriptsize [0.700,0.792] \\
Qwen3-0.6B & verses & 0.333 \scriptsize [0.275,0.392] & 0.527 \scriptsize [0.465,0.585] & 0.429 \scriptsize [0.377,0.481] & 0.463 \scriptsize [0.411,0.516] \\
Qwen3-0.6B & AI expl. & 0.624 \scriptsize [0.566,0.682] & 0.795 \scriptsize [0.744,0.845] & 0.696 \scriptsize [0.644,0.745] & 0.723 \scriptsize [0.673,0.771] \\
LaBSE & verses & 0.209 \scriptsize [0.163,0.260] & 0.368 \scriptsize [0.310,0.426] & 0.285 \scriptsize [0.237,0.333] & 0.312 \scriptsize [0.263,0.362] \\
LaBSE & AI expl. & 0.519 \scriptsize [0.457,0.581] & 0.670 \scriptsize [0.612,0.729] & 0.592 \scriptsize [0.537,0.644] & 0.620 \scriptsize [0.567,0.671] \\
Muffakir-v2 & verses & 0.132 \scriptsize [0.093,0.174] & 0.233 \scriptsize [0.182,0.287] & 0.190 \scriptsize [0.149,0.233] & 0.206 \scriptsize [0.163,0.251] \\
Muffakir-v2 & AI expl. & 0.442 \scriptsize [0.384,0.500] & 0.585 \scriptsize [0.527,0.647] & 0.508 \scriptsize [0.452,0.564] & 0.533 \scriptsize [0.478,0.588] \\
\textbf{BM25} & verses & 0.248 \scriptsize [0.198,0.302] & 0.392 \scriptsize [0.333,0.454] & 0.321 \scriptsize [0.272,0.372] & 0.349 \scriptsize [0.298,0.401] \\
\textbf{BM25} & AI expl. & 0.612 \scriptsize [0.554,0.670] & 0.760 \scriptsize [0.709,0.810] & 0.676 \scriptsize [0.624,0.726] & 0.698 \scriptsize [0.648,0.747] \\
\bottomrule
\end{tabular}

\caption{Track A in full, with bootstrap 95\% confidence intervals.}
\label{tab:apptracka}
\end{table*}

\section{The Second and Third Judges}
\label{app:judge2}
Track B was repeated with Gemma-3-27B-it, quantised to 4 bits and run on one consumer GPU, over a random $100$ of the $200$ thematic queries for all ten systems and both corpora ($2{,}000$ cells, no parse failures). The rubric text is identical to the primary judge's; only the response format differs, in that the second judge returns the seven scores with a single short justification rather than seven, which keeps local decoding tractable. Table~\ref{tab:appjudge2} gives system means under both judges and Table~\ref{tab:appjudge2rho} the cell-level agreement per aspect.

Agreement is high on the five aspects that ask about aboutness ($\rho$ between $0.65$ and $0.71$) and much lower on metaphor understanding ($0.26$) and literal match ($0.29$). The two aspects on which the judges agree least are the two that ask for the most interpretive judgment, and conclusions that rest on them are correspondingly weaker. The overall-quality ordering, which carries the Track B claim, is not one of them.

The third judge, Qwen3-32B quantised to 4 bits, scores the same 100 queries twice: once with the original rubric, and once with a neutral rubric that removes the literal-match aspect and its sentence about implicit meaning and changes nothing else. Table~\ref{tab:appjudges} gives system means on the shared queries under all judges.

\begin{table*}[t]
\centering\small
\setlength{\tabcolsep}{4pt}
\begin{tabular}{l cccc c cccc}
\toprule
& \multicolumn{4}{c}{\textit{FannOrFlop}} & & \multicolumn{4}{c}{\textit{Diwan}} \\
\cmidrule(lr){2-5}\cmidrule(lr){7-10}
System & Gem.\ & Gem-3 & Q-orig & Q-neut & & Gem.\ & Gem-3 & Q-orig & Q-neut \\
\midrule
Qwen3-8B & 2.83 & 3.36 & 3.17 & 3.35 & & 2.66 & 3.29 & 3.12 & 3.32 \\
BGE-M3 & 2.49 & 3.09 & 2.80 & 2.99 & & 2.33 & 3.00 & 2.77 & 2.98 \\
mE5-large & 2.48 & 3.23 & 3.03 & 3.11 & & 2.36 & 3.18 & 2.91 & 3.06 \\
jina-v3 & 2.51 & 3.19 & 2.86 & 3.10 & & 2.48 & 3.18 & 3.00 & 3.11 \\
silma-sts & 2.73 & 3.23 & 2.99 & 3.24 & & 2.55 & 3.17 & 3.00 & 3.28 \\
gte & 2.45 & 3.04 & 2.71 & 2.90 & & 2.10 & 2.90 & 2.59 & 2.75 \\
Qwen3-0.6B & 2.32 & 3.08 & 2.86 & 3.00 & & 2.18 & 3.02 & 2.59 & 2.73 \\
LaBSE & 1.93 & 2.82 & 2.32 & 2.50 & & 1.59 & 2.58 & 2.07 & 2.29 \\
Muffakir-v2 & 2.37 & 3.01 & 2.79 & 2.91 & & 2.16 & 2.99 & 2.81 & 2.98 \\
\midrule
\textbf{BM25} & 1.96 & 2.84 & 2.41 & 2.44 & & 1.69 & 2.69 & 2.30 & 2.46 \\
\midrule
$\tau$ with Gemini & ref. & 0.81 & 0.67 & 0.69 & & ref. & 0.85 & 0.89 & 0.91 \\
\bottomrule
\end{tabular}

\caption{Track B overall quality on the 100 thematic queries scored by every judge, verse mode: Gemini 3.5 Flash (Gem.), Gemma-3-27B-it (Gem-3), and Qwen3-32B under the original (Q-orig) and neutral (Q-neut) rubrics. The last row gives Kendall $\tau$ of each system ordering with Gemini's.}
\label{tab:appjudges}
\end{table*}

\begin{table*}[!t]
\centering\small
\begin{tabular}{llccccccc}
\toprule
Corpus & System & sem.\,rel. & them.\,coh. & metaph. & coverage & lit.\,match & prec. & overall \\
\midrule
FoF & Qwen3-8B & 3.00 & 2.70 & 2.94 & 3.08 & 3.27 & 2.80 & 2.94 \\
FoF & BGE-M3 & 2.54 & 2.33 & 2.55 & 2.58 & 3.25 & 2.37 & 2.47 \\
FoF & mE5-large & 2.60 & 2.46 & 2.66 & 2.58 & 3.28 & 2.37 & 2.52 \\
FoF & jina-v3 & 2.69 & 2.48 & 2.73 & 2.75 & 3.15 & 2.52 & 2.64 \\
FoF & silma-sts & 2.85 & 2.57 & 2.84 & 2.85 & 2.90 & 2.62 & 2.75 \\
FoF & gte & 2.42 & 2.21 & 2.38 & 2.46 & 3.40 & 2.22 & 2.37 \\
FoF & Qwen3-0.6B & 2.47 & 2.21 & 2.35 & 2.41 & 3.47 & 2.25 & 2.35 \\
FoF & LaBSE & 2.15 & 1.90 & 2.15 & 2.10 & 3.35 & 1.98 & 2.07 \\
FoF & Muffakir-v2 & 2.48 & 2.31 & 2.46 & 2.48 & 3.17 & 2.33 & 2.43 \\
FoF & \textbf{BM25} & 1.95 & 1.73 & 1.87 & 1.94 & 4.26 & 1.78 & 1.90 \\
\midrule
Diwan & Qwen3-8B & 3.00 & 2.65 & 2.91 & 2.89 & 3.29 & 2.77 & 2.83 \\
Diwan & BGE-M3 & 2.42 & 2.27 & 2.45 & 2.41 & 3.38 & 2.24 & 2.29 \\
Diwan & mE5-large & 2.44 & 2.29 & 2.45 & 2.40 & 3.40 & 2.21 & 2.35 \\
Diwan & jina-v3 & 2.65 & 2.44 & 2.68 & 2.60 & 3.29 & 2.45 & 2.53 \\
Diwan & silma-sts & 2.75 & 2.50 & 2.70 & 2.73 & 3.05 & 2.52 & 2.63 \\
Diwan & gte & 2.27 & 2.06 & 2.27 & 2.26 & 3.54 & 2.10 & 2.18 \\
Diwan & Qwen3-0.6B & 2.23 & 2.03 & 2.12 & 2.15 & 3.71 & 2.04 & 2.14 \\
Diwan & LaBSE & 1.86 & 1.65 & 1.82 & 1.77 & 3.76 & 1.62 & 1.75 \\
Diwan & Muffakir-v2 & 2.41 & 2.19 & 2.41 & 2.29 & 3.23 & 2.16 & 2.27 \\
Diwan & \textbf{BM25} & 1.81 & 1.64 & 1.69 & 1.74 & 4.48 & 1.65 & 1.70 \\
\bottomrule
\end{tabular}

\caption{Track B by judge aspect, verse mode, $n{=}200$ queries per cell. All aspects are scored from 1 to 5 with 5 highest; on \textit{lit.\ match} a high score means the results repeat the query's wording, which the rubric treats as evidence against implicit understanding. Lexical overlap can coexist with genuine relevance, and no claim in the paper rests on this aspect: the ordering from the aboutness aspects alone matches the overall one.}
\label{tab:appaspects}
\end{table*}

\begin{table}[h]
\centering\small
\setlength{\tabcolsep}{4pt}
\begin{tabular}{l cc c cc}
\toprule
& \multicolumn{2}{c}{\textit{FannOrFlop}} & & \multicolumn{2}{c}{\textit{Diwan}} \\
\cmidrule(lr){2-3}\cmidrule(lr){5-6}
System & judge 1 & judge 2 & & judge 1 & judge 2 \\
\midrule
Qwen3-8B & 2.83 & 3.36 & & 2.68 & 3.29 \\
BGE-M3 & 2.49 & 3.09 & & 2.34 & 3.00 \\
mE5-large & 2.48 & 3.23 & & 2.36 & 3.18 \\
jina-v3 & 2.51 & 3.19 & & 2.48 & 3.18 \\
silma-sts & 2.73 & 3.23 & & 2.56 & 3.17 \\
gte & 2.45 & 3.04 & & 2.10 & 2.90 \\
Qwen3-0.6B & 2.32 & 3.08 & & 2.18 & 3.02 \\
LaBSE & 1.93 & 2.82 & & 1.59 & 2.58 \\
Muffakir-v2 & 2.37 & 3.02 & & 2.16 & 2.99 \\
\midrule
\textbf{BM25} & 1.96 & 2.84 & & 1.70 & 2.69 \\
\bottomrule
\end{tabular}

\caption{Track B overall quality under the primary judge (Gemini 3.5 Flash, $n{=}200$) and the open-weights second judge (Gemma-3-27B-it, $n{=}100$). Scales differ; the orderings do not (Kendall $\tau{=}{+}0.81$ on \textit{Fann or Flop}, ${+}0.85$ on \textit{Diwan}).}
\label{tab:appjudge2}
\end{table}

\begin{table}[t!]
\centering\small
\begin{tabular}{lc}
\toprule
Aspect & cell-level $\rho$ \\
\midrule
semantic relevance & +0.71 \\
thematic coherence & +0.64 \\
metaphor understanding & +0.26 \\
coverage & +0.68 \\
literal match & +0.29 \\
precision & +0.65 \\
overall quality & +0.67 \\
\bottomrule
\end{tabular}

\caption{Cell-level Spearman agreement between the two judges, $n{=}2{,}000$ paired cells.}
\label{tab:appjudge2rho}
\end{table}

\section{Track B by Judge Aspect}
\label{app:aspects}
Table~\ref{tab:appaspects} breaks Track B out by all seven rubric aspects. Two patterns stand out. First, BM25 is last of ten on every aspect that requires aboutness on \textit{Fann or Flop} and on all but one on \textit{Diwan}, with its largest deficits on metaphor understanding and semantic coverage ($1.1$ to $1.2$ points below the best system), which is where an aboutness-blind system should fail. Second, on \emph{direct textual relevance}, the aspect that measures how far a system's results simply repeat the query's wording, BM25 scores highest of all ten systems on both corpora ($4.26$ on \textit{Fann or Flop}, $4.48$ on \textit{Diwan}). The judge therefore identifies BM25 as the most surface-matching system while scoring it lowest overall. This agrees with the mechanism of \S\ref{sec:bridge} through a separate measurement, and shows that the judge's low overall score for BM25 is not a blanket penalty on lexical retrieval.

\section{Human Annotation}
\label{app:human}
\textbf{Thematic relevance.} Twenty thematic queries, ten per corpus, were pooled across three systems (BM25, jina-v3 and their reciprocal-rank fusion), top 3 each, deduplicated to 126 items and shuffled with system identity removed. Two authors graded every item independently on the 0 to 2 scale of Appendix~\ref{app:annot}. They agree moderately: quadratic-weighted $\kappa{=}0.53$, with $90\%$ of items within one point. The second rater is more lenient (mean $0.90$ against $0.68$), but both rank the systems the same way, with BM25 below jina-v3 for each rater ($p{=}.005$ and $p{=}.028$, paired Wilcoxon), and both agree with the judge's overall-quality score on the $40$ shared cells ($\rho{=}0.75$ and $0.64$).

\textbf{Known-item spot check.} For $50$ queries where Qwen3-8B, the strongest dense encoder, missed the gold poem, one author judged whether its top-1 result would satisfy the query. The result was a fully acceptable answer in $34\%$ of cases (95\% CI $[22,48]$) and at least partly acceptable in $72\%$. Twelve control items showing the gold poem itself, shuffled among the rest, were judged fully acceptable in $11$ of $12$, so the annotator was discriminating rather than lenient. The rate is higher on the abstract corpus ($0.40$) than on the tip-of-the-tongue one ($0.28$). Applying each corpus its own rate, Qwen3-8B's Recall@5 rises from $0.717$ to $0.813$ on \textit{Fann or Flop} and from $0.748$ to $0.834$ on \textit{Diwan}, and the quotation interaction of \S\ref{sec:strat} shrinks from $+0.16$ to $+0.11$.

\section{Corpus Checks}
\label{app:corpus}
\textbf{Where the quoted spans sit.} Of the $80$ quoted spans in \textit{Fann or Flop}, $37$ come from the opening verse, $41$ from a middle verse and $2$ from the final verse, and $77$ occur in no other poem of the corpus. BM25 retrieves the gold poem for all of them. The edits of \S\ref{sec:edits} vary the position directly.

\textbf{Era.} On \textit{Fann or Flop}, we group the era labels into four ordered bins and fit a logistic model of Recall@5 success that controls for query-verse containment and query length. Pooled over the nine dense encoders (system fixed effects, errors clustered by query), success rises by $+0.27$ log-odds per step towards the present ($p{=}.010$). BM25 shows no such trend ($-0.36$, $p{=}.14$), and the difference between the two is significant ($p{=}.015$).

\textbf{Poet frequency.} As a rough proxy for exposure during pretraining, the number of poems a poet has in \textit{Fann or Flop} (from $1$ to $612$) is unrelated to retrieval success for every system ($|\rho|\leq.08$, all $p>.2$). \textit{Diwan} cannot test this, since every poet contributes $70$ poems.

\textbf{Target selection.} The gold poems do not differ from the rest of their corpus in era, genre or length on \textit{Fann or Flop}, nor in era, genre, length or poet on \textit{Diwan} (chi-square and Mann-Whitney tests, all $p>.05$).

\textbf{Duplicates.} In \textit{Diwan}, $274$ of the $3{,}500$ poems share their normalised text with another poem ($135$ groups), and $28$ gold poems have such a twin. Counting a twin of the gold poem as correct changes verse-mode Recall@5 by $0.4$ points for one system and leaves the other nine unchanged. \textit{Fann or Flop} has no duplicates.

\textbf{Explanation audit.} A co-author rated $50$ random verse explanations against their verses. Of these, $60\%$ were faithful, $32\%$ had a minor error and $8\%$ misread the verse; $20\%$ added substantive content not in the verse. The failure modes were inverted meaning, invented names and classical vocabulary read in its modern sense.

\section{Annotation Guidelines}
\label{app:annot}
\textbf{Thematic relevance.} Annotators saw a theme and a shuffled, deduplicated pool of poems retrieved for it by several anonymised systems, and graded each poem: 2 = clearly on-theme, 1 = partly or tangentially on-theme, 0 = off-theme. The instructions stated explicitly that lexical overlap with the theme is not evidence of relevance, that a poem repeating the theme's words while being about something else scores 0, and that a poem never using them while genuinely being about the theme scores 2.

\textbf{Known-item spot check.} Each item showed one query and one retrieved poem, and asked only whether that poem would satisfy someone issuing the query: 2 = yes, 1 = partly or arguably, 0 = no. The instructions stated that the annotator would not be told which items showed the labelled gold poem and that guessing was counter-productive, and that a query may legitimately describe more than one poem, so a genuine answer scores 2 even if a different poem was intended. Both guideline documents and the completed judgments are released.

\section{LLM Prompts and Rubric}
\label{app:prompts}
\textbf{Explanation generation.} Each verse is paired with a fixed instruction asking for a concise, faithful prose paraphrase of the verse's meaning in Arabic, adding no information beyond the verse.

\textbf{Judge rubric.} The judge receives a query and the retrieved poems, each given as its first six verses, and scores seven aspects from 1 to 5, each with a one-line justification: semantic relevance, thematic coherence, metaphor understanding, semantic coverage, direct textual relevance (where 1 is best, i.e.\ least literal matching), precision, and overall quality. The neutral rubric used with the third judge drops direct textual relevance. Exact prompt strings are released with the code.

\end{document}